\pdfoutput=1

\documentclass[11pt]{article}

\usepackage[]{acl}

\usepackage{times}
\usepackage{latexsym}
\usepackage{graphicx}
\usepackage[T1]{fontenc}

\usepackage[utf8]{inputenc}
\usepackage{multirow}  

\usepackage{algorithm2e}

\usepackage{microtype}

\title{Reevaluating Data Partitioning for Emotion Detection in EmoWOZ}

\author{Moeen Mostafavi \\
  Systems and Information Engineering\\
University of Virginia\\
Charlottesville, VA, USA\\
  \texttt{moeen@virginia.edu} \\\And
  Michael D. Porter   \\
  Data Science \& \\ Systems and Information Engineering   \\
University of Virginia\\
Charlottesville, VA, USA \\
  \texttt{mdp2u@virginia.edu} \\}

\begin{document}
\maketitle
\begin{abstract}
This paper focuses on the EmoWoz dataset, an extension of MultiWOZ that provides emotion labels for the dialogues. MultiWOZ was partitioned initially for another purpose, resulting in a distributional shift when considering the new purpose of emotion recognition. The emotion tags in EmoWoz are highly imbalanced and unevenly distributed across the partitions, which causes sub-optimal performance and poor comparison of models. We propose a stratified sampling scheme based on emotion tags to address this issue, improve the dataset's distribution, and reduce dataset shift. We also introduce a special technique to handle conversation (sequential) data with many emotional tags. Using our proposed sampling method, models built upon EmoWoz can perform better, making it a more reliable resource for training conversational agents with emotional intelligence. We recommend that future researchers use this new partitioning to ensure consistent and accurate performance evaluations.
\end{abstract}

\section{Introduction}

Emotion recognition in task-oriented conversational agents is challenging because it requires the agent to accurately interpret and respond to a user's emotional state in real time. Emotional signals can be complex and difficult to detect accurately, especially in unstructured conversations where users may not express their emotions explicitly. Another challenge is that emotions can be context-dependent, meaning that they may be influenced by the user's past experiences, current environment, and cultural background. Therefore, conversational agents need to interpret these contextual factors accurately to provide appropriate responses sensitive to the user's emotional state.

Despite the challenges, emotion recognition in task-oriented conversational agents is important because it can improve the overall user experience \cite{zhang2020towards}. By accurately detecting and responding to a user's emotional state, conversational agents can provide more personalized and empathetic interactions, increasing user satisfaction and engagement. Additionally, emotion recognition can help agents identify when a user is experiencing frustration, confusion, or other negative emotions, allowing them to intervene and provide support to prevent user dropout or dissatisfaction\cite{andre2004endowing}. Emotion recognition is critical to developing effective task-oriented conversational agents that can provide a human-like user experience \cite{polzin2000emotion}.

Sentiment analysis in task-oriented conversational agents has been addressed in the literature as an essential aspect of natural language processing that can improve the overall user experience \cite{shi2018sentiment, saha2020towards, wang2020sentiment}.
However, the lack of publicly available data for emotion recognition is a significant limitation for task-oriented conversational agent applications. 

MultiWOZ (Multi-Domain Wizard-of-Oz) is a large-scale dataset of human-human written conversations for task-oriented dialogue modeling. The dataset was initially collected for training and evaluating dialogue systems, particularly those designed to assist users with completing specific tasks such as booking a hotel or reserving a table at a restaurant \cite{budzianowski2018multiwoz}. \citet{feng2021emowoz} extended the MultiWOZ dataset by including dialogues between humans and a machine-generated policy, which they named DialMAGE. The resulting dataset was called EmoWOZ.  EmoWOZ is a large-scale, manually emotion-annotated corpus of task-oriented dialogues. The corpus contains more than 11K dialogues with more than 83K emotion annotations of user utterances, which makes it the first large-scale open-source corpus of its kind. The authors propose a novel emotion labeling scheme tailored to task-oriented dialogues and demonstrate the usability of this corpus for emotion recognition and state tracking in task-oriented dialogues. The paper highlights that while emotions in chit-chat dialogues have received considerable attention \cite{li2017dailydialog, poria2018meld, zahiri2017emotion}, emotions in task-oriented dialogues remain largely unaddressed. They argue that incorporating emotional intelligence can help conversational AI generate more emotionally and semantically appropriate responses, making a better user experience.

In their study, the authors described their methodology for partitioning the EmoWOZ dataset into training, validation, and testing sets while maintaining the original split of the MultiWOZ dataset. They also divided the DialMAGE dataset into three parts with a ratio of 8:1:1. However, in the following section, we will examine the limitations of their approach to data partitioning and suggest an alternative method.

We specifically concentrate on the MultiWOZ section of EmoWOZ in this paper as it is widely used in various applications. Therefore, we do not discuss all subsets of the EmoWOZ in detail. Nonetheless, our approach can be readily extended to the entire EmoWOZ dataset.

\section{Data partitioning}
When building a predictive model, we typically split our data into three sets: a training set, a validation set, and a test set. The purpose of the training set is to estimate model parameters, and the purpose of the validation set is to tune the model's hyperparameters and assess its performance. The test set aims to get an unbiased estimate of the model's performance on new, unseen data.
It's important to note that the test set should not be used in any part of model fitting or model selection because doing so can lead to overfitting and inaccurate performance estimate. However, after we have trained and validated our models on the training and validation sets, we can use the test set to compare the performance of different models and select the best one.
This is where the issue of different distributions between the test and validation sets comes in. If the test set has a different distribution than the validation set, a model that performs well on the validation set may not be the best model for the test set, and vice versa. 
Therefore, it's essential to ensure that the distributions of the three sets are as similar as possible. 
\begin{table}[]
\setlength{\tabcolsep}{4pt}
\small
\begin{tabular}{c|ccc||ccc}
           & \multicolumn{3}{c}{Relative Frequency}
            
           & \multicolumn{3}{c}{Manual   resolution} \\   
           & Fear.   & Abus.   & Dis.   & Fear.    & Abus.   & Dis.   \\ \hline
Train      & 0.62      & 0.06      & 1.29           & 28.86      & 87.88     & 16.53          \\
Devel. & 0.22      & 0.08      & 1.00           & 62.50       & 50.00      & 21.62          \\
Test       & 0.20      & 0.07      & 1.47           & 40.00      & 100.0       & 20.37        
\end{tabular}
\caption{\label{distribution}
Relative frequency and Manual resolution percentage for the three minority classes with less than 2\% relative frequency in the MultiWOZ dataset} 
\end{table}

Approximately 2.5\% of the MultiWOZ subset in \cite{feng2021emowoz} underwent manual annotation for resolution. However, this manual annotation was not evenly distributed across all classes, and minority classes were affected more than majority classes. For instance, the \textit{abusive} class in the test set was completely annotated manually, while in the developing set it was manually labeled in 50\% of cases. Also, the \textit{Fearful} class had a relative frequency three times higher in the training set, as shown in Table~\ref{distribution}.

Three annotators labeled the utterances according to the task. The final label was determined primarily by the majority vote of the annotators. Among all utterances, 72.1\% had a complete agreement among the three annotators. A partial agreement was found for 26.4\% of the utterances, while for 1.5\%, there was no agreement. The paper reports that these instances were resolved manually to address cases where the annotators could not reach an agreement. In a small portion of the data, a label different from the majority vote was chosen. 

We use F1 scores to compare annotators across data partitions to assess inter-annotator agreement. In essence, we measure the effectiveness of annotations by the three annotators across the training, validation, and test sets using the final labels. Table~\ref{annotator_per} presents the results, indicating model performance discrepancies across the training, development, and test sets.

\begin{table}
\setlength{\tabcolsep}{4pt}
\small
 \setlength{\tabcolsep}{1pt}
    \begin{tabular}{c|c|c|c|c|c|c|c||c|c}
\multirow{2}{*}{ Data } & \multicolumn{7}{c}{F1 for each Emotion Label} & \multicolumn{2}{c}{Macro F1 }  \\
& Neu. & Fea. & Dis. & Apo. & Abu. & Exc. & Sat.  & w/o N.  & w N. \\  \hline
Train & 93.94 & 54.71 & 50.19 & 72.85 & 32.49 & 42.32 & 88.78 & 56.89 &   62.18  \\
Val.& 94.09 & 26.25 & 46.72 & 73.35 & 50.00 & 43.19 & 88.73 & 54.71  & 60.33  \\
Test & 94.14 & 29.73 & 52.03 & 71.98 & 32.00 & 34.97 & 88.86 & 51.6 &  57.67  \\ \hline
\end{tabular} 
\caption{\label{annotator_per}
Performance of annotators based on F1 for emotion labels (\textbf{Neu}tral, \textbf{Fea}rful, \textbf{Dis}satisfied, \textbf{Apo}logetic, \textbf{Abu}sive, \textbf{Exc}ited, \textbf{Sat}isfied) on MultiWOZ}. Following the benchmarks in the literature, in the aggregated level, we report Macro F1 scores \textbf{w}ith\textbf{o}out \textbf{N}eutral and \textbf{w}ith \textbf{N}eutral emotion.  
\end{table}
  This phenomenon is known as \textit{dataset shift} in which there is a difference between the joint distribution of inputs and outputs during the training stage compared to the validation and test stage \cite{quinonero2008dataset}, leading to a decrease in performance.
Dataset Shift is a common problem in machine learning, and it can have significant consequences, such as a decrease in accuracy. In the case of EmoWoz, the dataset Shift arises from the fact that \citep{feng2021emowoz} kept the original partitioning of MultiWOZ, which is not evenly distributed across partitions for this particular task. 

It's worth noting that while the original partitioning of MultiWOZ data is suitable for many tasks related to the development of task-oriented conversational agents, it may not be ideal for emotion detection. Emotion detection requires consideration of different contextual aspects, which may require a new partitioning approach. For instance, in the original partitioning, 60\% of messages with the \textit{Fearful} emotion in the training set were in conversations with the police or hospital. However, in the validation and test sets, none of the conversations with \textit{Fearful} emotions were related to the police or hospital. This contextual aspect of the conversation plays a crucial role in accurately recognizing emotions.

To address this issue, we use stratified sampling, which is a sampling technique that ensures that each sub-group in the data is represented proportionally in the sample. In this case, we use stratified sampling to ensure that the training set has a distribution similar to the validation and test sets.
Stratified sampling is particularly useful in situations where the distribution of the target variable is imbalanced or varies across sub-groups in the data. In the case of EmoWoz, the distribution of emotions in the training set has differed from that in the validation and test sets, which could have contributed to the dataset shift. We used the Algorithm \ref{alg:two} to get a new partitioning of the data. 
\begin{algorithm}[hbt!]
\caption{Stratified sampling for emotion recognition in the conversation}\label{alg:two}
\KwData{MultiWoz dataset with emotion labels from EmoWOZ}
\begin{enumerate}
    \item Group the dataset based on their utterance ids and find a list with the emotion sequence in each utterance.
    \item  Determine the frequency of emotional sequences in the dataset.
\item Make a dictionary called \textit{emotion\_seq\_dict} with the emotional sequence as the key and the counts of the sequence in the dataset as the value.
\item Partition the whole dataset into one set called \textit{frequent\_seq} with conversations of more than six emotional list frequencies and another set \textit{non\_frequent\_seq} with the rest of the data.
\item For the \textit{frequent\_seq}, do the stratified sampling of conversation based on the emotion sequence and partition it to the training, validation, and test set, with a 80-10-10 split similar to the original split of the data.
\item Use random sampling to partition the \textit{non\_frequent\_seq} to the training, validation, and test sets.
\item Find the union of the two partitions to get the partitioning of the whole dataset.
\end{enumerate}
\end{algorithm}
By using stratified sampling, we have ensured that the emotion distribution in the training set was similar to that in the validation and test sets, which helps to reduce the dataset Shift and improve the model's performance. Table~\ref{annotator_per2} shows the annotator's F1-score after the new partitioning.

\begin{table}
\setlength{\tabcolsep}{4pt}
\small
 \setlength{\tabcolsep}{1pt}
    \begin{tabular}{c|c|c|c|c|c|c|c||c|c}
\multirow{2}{*}{ Data } & \multicolumn{7}{c}{F1 for each Emotion Label} & \multicolumn{2}{c}{Macro F1 }  \\
& Neu. & Fea. & Dis. & Apo. & Abu. & Exc. & Sat.  & w/o N.  & w N. \\  
  \hline
Train & 94.02 & 52.38 & 50.97 & 73.06 & 34.87 & 41.73 & 88.83 & 56.98 &  62.27  \\
Val. & 93.86 & 48.58 & 47.16 & 71.49 & 37.50 & 41.65 & 88.69 & 55.85 &  61.28  \\
Test & 93.79 & 52.08 & 46.13 & 72.20 & 32.26 & 41.54 & 88.49 & 55.45 &  60.93  \\ \hline
\end{tabular} 
\caption{\label{annotator_per2}
Performance of annotators based on F1 for emotion labels (\textbf{Neu}tral, \textbf{Fea}rful, \textbf{Dis}satisfied, \textbf{Apo}logetic, \textbf{Abu}sive, \textbf{Exc}ited, \textbf{Sat}isfied) on MultiWOZ after new partitioning.} 
\end{table}

\section{Case study}
\citet{feng2021emowoz} used Bert, Contextual BERT, DialogueRNN \cite{majumder2019dialoguernn}, and COSMIC \cite{ghosal2020cosmic} for the baseline methods. Among these methods, BERT did not incorporate the sequential aspects of the conversation, yet it yielded the best Macro F1 scores in most EmoWOZ subsets. To enhance the BERT model, we compute the relative embedding of the message from the chatbot and agent and employ a transformer model to address the sequential aspects of the problem. Hyperparameters for this method using both the original and proposed partitioning are illustrated in Table~\ref{TSPD_per}. Upon examining the results, we can see that in the original partitioning, the hyperparameters corresponding to the top-performing model on the validation set produced the worst model on the test set. This discrepancy could be indicative of data drift, as we previously discussed. 
\begin{table}[]
\small
\setlength{\tabcolsep}{2.5pt}
\begin{tabular}{cc|ccc||ccc}
      &        & \multicolumn{3}{c}{Original splits}   & \multicolumn{3}{c}{Stratified splits}        \\
      \hline
Batch & Epochs & Val.           & Test & Dif.          & Val.           & Test   & Dif.        \\
\hline
8     & 4      & \textbf{51.4} & 48.92 & -2.48
         & 52.25         & 51.58     & -0.67     \\
16    & 4      & 50.58         & 54.03  & 3.45         & 53.09         & 51.76  & -1.33        \\
32    & 4      & 48.96         & \textbf{54.23} & 5.27
 & 55            & \textbf{53.73} & -1.27  \\
8     & 8      & 49.52         & 52.35 & 2.83         & 53.5          & 52.38  & -1.12        \\
16    & 8      & 50.31         & 51.77   & 1.46       & 53.54         & 49.68  & -3.86        \\
32    & 8      & 49.86 
        & 52.27  & 2.41        & \textbf{56.8} & 53.14  & -3.66
       
\end{tabular}
\caption{\label{TSPD_per}
Performance of different hyper-parameters.}
\end{table}
\begin{figure}[h]
  \centering
  \includegraphics[width=0.4\textwidth,trim={0cm 8.4cm 3cm 0},clip]{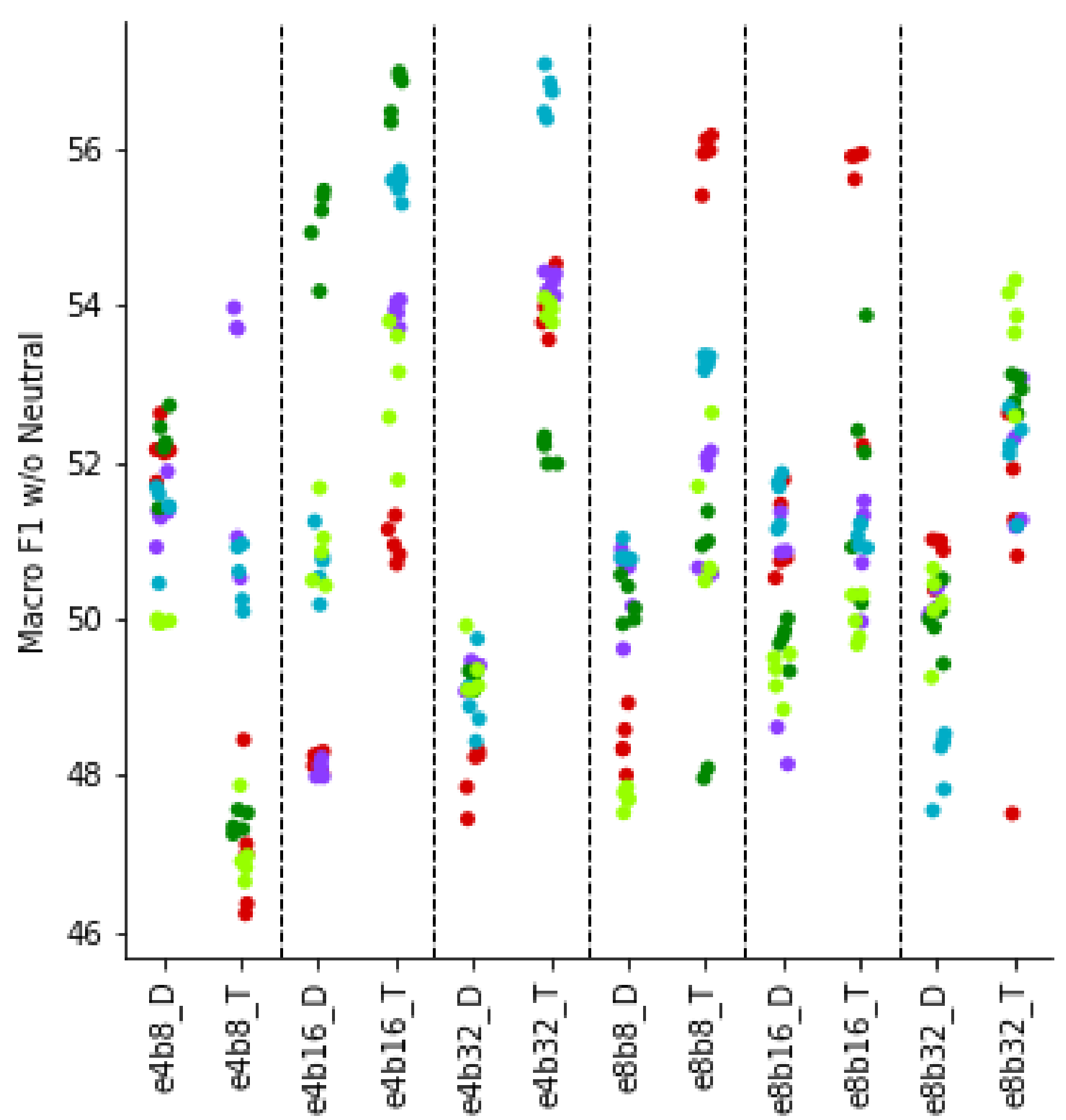}
  \caption{This figure depicts the Macro F1 score for each of the seeds utilized in implementing the sequential extension of the BERT model on the \textbf{origial} Multiwoz partitioning. For each hyperparameter, both the Macro F1 score in the validation and test sets are plotted in close proximity to one another. Additionally, the colors within the figure represent the five distinct seeds utilized in the embedding step.}
  \label{fig:cropped-example}
\end{figure}
\begin{figure}[h]
  \centering
  \includegraphics[width=0.4\textwidth,trim={0cm 8.4cm 3cm 0},clip]{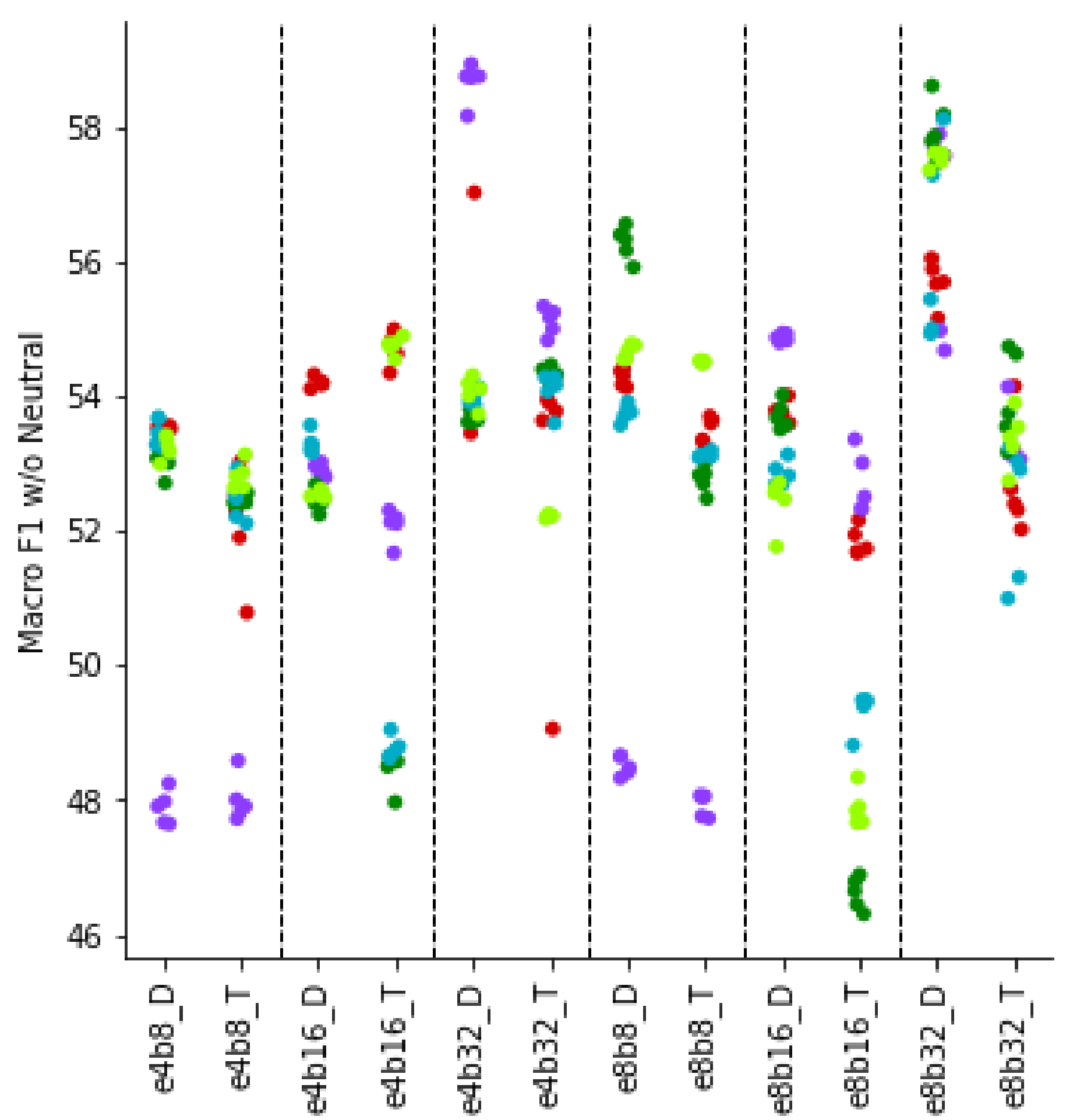}
  \caption{This figure depicts the Macro F1 score for each of the seeds utilized in implementing the sequential extension of the BERT model on the \textbf{proposed} Multiwoz partitioning. For each hyperparameter, both the Macro F1 score in the validation and test sets are plotted in close proximity to one another. Additionally, the colors within the figure represent the five distinct seeds utilized in the embedding step.}
  \label{fig:partition}
\end{figure}
This implementation uses five distinct seeds to generate five unique embeddings. We then employed five seeds to construct the transformer model on top of these embeddings. Consequently, we had 25 different models for each parameter in Table~\ref{TSPD_per}. To visualize the Macro F1 score for each of these models, we included Figures \ref{fig:cropped-example} and \ref{fig:partition}. The colors within these figures correspond to the five distinct seeds utilized in the embedding step. Notably, we can observe that only in the proposed partitioning of the data the change in averaged Macro F1 score is similar across the validation and test sets for various hyperparameters. Furthermore, we can observe that for models with equivalent hyperparameters, the change in score across different initial seeds is comparable in both the development and test sets. These observations suggest that no data drift is present in the new partitioning.
\section{Concluding remarks}
After analyzing the original data partitioning, we have identified potential data drift and suggested an alternative approach to address this issue. Our evaluation of the results indicates that the new partitioning approach effectively reduces data drift, as demonstrated by the consistency of Macro F1 scores in both the validation and test sets across different hyperparameters and initial seeds. These findings suggest that the proposed partitioning method is a suitable alternative for researchers working on emotion detection using MultiWOZ data. This work emphasizes the significance of meticulously choosing and employing partitioning methods in the training and assessment of machine learning models.

\bibliography{anthology,custom}
\bibliographystyle{acl_natbib}

\end{document}